\documentclass[11pt,twoside,twocolumn,a4paper]{article}

\usepackage{cvww}
\usepackage{times}
\usepackage{epsfig}
\usepackage{graphicx}
\usepackage{amsmath}
\usepackage{amssymb}
\usepackage{subcaption}
\usepackage{tabularx}
\usepackage{setspace}

\usepackage[pagebackref=true,breaklinks=true,bookmarks=false]{hyperref}
\cvwwfinalcopy % *** Uncomment this line for the final submission

\def\cvwwPaperID{27} % *** Enter the CVWW Paper ID here

\ifcvwwfinal\fi

\newcommand{\TODO}[1]{}
\renewcommand{\TODO}[1]{{\color{red} TODO: {#1}}}
\begin{document}

%%%%%%%%% TITLE
%\title{Noise Evaluation in Data Captured by 3D Cameras}
%\title{Enhancement of Synthetic Training Data using Noise Evaluation}
\title{Enhancement of 3D Camera Synthetic Training Data with Noise Models}
%\title{Towards Modelling of Noise in Synthetic 3D Scans}
%\title{Concept of Noise Synthetic Data Enhancement}
%\title{Enhancement of Synthetic Training Data based on Noise Evaluation}

%\author{Katarína Osvaldová, Lukáš Gajdošech, Viktor Kocur\\
%Department of Applied Informatics, Comenius University 
%Bratislava, Slovakia\\
%{\tt\small lukas.gajdosech@fmph.uniba.sk, viktor.kocur@fmph.uniba.sk}
%\and
%Martin Madaras \\
%Department of Applied Informatics, Comenius University Bratislava, Slovakia\\
%Skeletex Research, Slovakia\\
%{\tt\small madaras@skeletex.xyz}
%}

 \author{
 Katarína Osvaldová$^1$ \quad Lukáš Gajdošech$^1$ \quad Viktor Kocur$^1$ \quad Martin Madaras$^{1,2}$\\
 $^1$Faculty of Mathematics, Physics and Informatics, Comenius University in
 Bratislava\\
 $^2$Skeletex Research, Slovakia\\
 {\tt\small lukas.gajdosech@fmph.uniba.sk, viktor.kocur@fmph.uniba.sk, madaras@skeletex.xyz}
 }

\maketitle
\ifcvwwfinal\thispagestyle{fancy}\fi

\begin{abstract}
%The aim of this paper is evaluation of noise in data captured by 3D cameras. For investigation of the noise, a dataset containing scenes 
%captured simultaneously by multiple 3D cameras was created. This paper offers an overview of the various types of noise present in these images.
The goal of this paper is to assess the impact of noise in 3D camera-captured data by modeling the noise of the imaging process and applying it on synthetic training data. We compiled a dataset of specifically constructed scenes to obtain a noise model. We specifically model lateral noise, affecting the position of captured points in the image plane, and axial noise, affecting the position along the axis perpendicular to the image plane. The estimated models can be used to emulate noise in synthetic training data. The added benefit of adding artificial noise is evaluated in an experiment with rendered data for object segmentation. We train a series of neural networks with varying levels of noise in the data and measure their ability to generalize on real data. The results show that using too little or too much noise can hurt the networks' performance indicating that obtaining a model of noise from real scanners is beneficial for synthetic data generation.

%for the purpose of introducing noise to artificially generated training data. %For investigation of the noise, a dataset containing scenes captured simultaneously by multiple 3D cameras was created. % with the help of a custom application able to communicate with the devices. 
%This paper offers an overview of the various types of noise present in these images. %captured by 3D cameras. 
%The most attention is paid to lateral noise, affecting the position of captured points in the image plane, and axial noise, affecting the position along the axis perpendicular to the image plane.
%Using the captured dataset, a probabilistic noise model was created, for the purpose of addition of noise to artificially created range images. 
%The added benefit of adding artificial noise is evaluated in an experiment with rendered synthetic data for object segmentation. We train a series of neural networks with varying noise amount and measure their ability to generalize on real data. %For the creation of noise models for various devices, a script was created that computes the noise model based on a series of range images capturing specific scenes. 
% The final section contains example range images with artificial noise. With the help of these range images, various possible improvements to the model were presented.
\end{abstract}

\section{Introduction}

In the past, 3D cameras were rare and expensive. Nowadays, a plethora of 3D cameras of various quality and price are commercially available. As is the case with any camera, range data captured by these devices suffer from the presence of noise. 

The intersection of machine learning and computer vision has emerged as a dynamic field with diverse applications. Notably, the synthesis of these domains has become increasingly prominent. Machine learning requires training data, manual creation of which is not only time-consuming but also expensive. The advent of computer graphics has facilitated the cost-effective generation of synthetic data. Nevertheless, synthetic data lacks the inherent noise present in real 3D camera-captured data, leading to a domain gap. To bridge this gap, the process of synthetic data creation may involve the intentional addition of artificial noise. %Crucially, understanding the nature of noise specific to a given device is essential to accurately emulate it. The overarching objective is to establish a robust pipeline for the generation of realistic synthetic data, mirroring the characteristics of data as if captured by a physical 3D camera device.
Noise is, however, a complex topic. Countless factors influence its behaviour, from the technology employed by the device, design and quality, through environmental variables, such as ambient light and temperature, to properties of the scene. The presence of some noise can be avoided, and in some cases, the noise can be modelled. 

Noise has been the topic of numerous studies~\cite{belhedi2015noise, chatterjee2015noise, falie2007noise, mallick2014characterizations, nguyen2012modeling}. Most of them, however, focus on investigation of one specific device or principle. Some works focus on theoretical models of noise. These models serve as a guide for investigation of the noise of specific devices, as the parameters of the devices needed for the employment of the models are usually not publicly available and are subject to trade secrets. 

%This work compiles an overview of noise types present in the measurements of cameras employing the most common technologies on the market. For each noise type, the cause of its presence is investigated. 
%Some types of noise present in the measurements of cameras employing the most common technologies can be modelled simply from their definition. Other require comprehensive analysis before model creation. 

Axial and lateral noise of 3D cameras were chosen for a comprehensive investigation, as the theoretical models describing their behaviour rely heavily on knowledge of publicly undisclosed parameters. We have collected a dataset of several thousands scans from three different devices to fit probabilistic models of noise with respect to the distance of the imaged objects and angles of their surface.

We also performed an experiment with a segmentation neural network trained on synthetic data. We varied the amount of noise added to the generated data. Evaluation on real scans shows that using too little or too much noise can hurt the network's performance. 
%as the best performance was achieved by adding slightly more noise than estimated from devices. 
The knowledge of noise parameters of real devices can thus be beneficial when employing synthetic data for training deep neural networks.

\section{Related Work}
\label{sec:related}

Various approaches have been explored to enhance the accuracy and efficiency of 3D scanning technologies, with a particular focus on training datasets that fuel machine learning models behind the processing pipelines. Understanding the artifacts inherent in scanning technologies is crucial for generating training data that accurately reflects the real world's variance. Different 3D scanning methods, such as structured light triangulation and time-of-flight measurements, introduce unique artifacts that can impact data quality. Some common artifacts include noise, distortions, and systematic errors.

\subsection{Structured Light Scanning}

Structured Light (SL) triangulation is based on the principles of two-view geometry. One camera is replaced by a light source that projects a sequence of patterns onto the scene. The patterns projected get deformed by the geometric shapes of the objects in the scene. A camera situated at a fixed distance from the projector then captures the scene with the projected pattern~\cite{sonka2014image}. By analysing the distortion of the pattern, information about position of the objects in the scene can be determined. %The principle is illustrated in the \autoref{SLIlustration}.

Various patterns have been proposed~\cite{sarbolandi2015kinect}. For example, the Kinect v1 camera uses a fixed dot pattern~\cite{kinectPattent}. Photoneo's MotionCam-3D camera utilises parallel structured light technology which enables the device to capture the scene depth at high resolution and frame-rate at the same time~\cite{photoNeoPattent}.

\subsection{Time-of-Flight Scanning}

Time-of-Flight (ToF) measurement technology is based on the principle of calculating the distance of an object in the scene by measuring the time it takes for an emitted signal to travel to the object and back. The distance is calculated from measurements of phase difference~\cite{hansard2012time}. 
The exact type of waves employed varies based on the application. 
RADAR and LIDAR include ToF measurements~\cite{sonka2014image}. The most common approach in ToF cameras is the continuous-wave intensity modulation IR LIDAR~\cite{gschwandtner2011blensor}. The distance is calculated from the observed phase delay of the amplitude envelope of the reflected light~\cite{tolgyessy2021evaluation}.
%Intensity\nobreakdash-modulated light source, most commonly LEDs, illuminate the whole scene at once \cite{gschwandtner2011blensor, sarbolandi2015kinect}. 

The range and accuracy of ToF devices are primarily influenced by the wavelength and energy of the emitted light, necessitating safety precautions, including energy capping in human environments~\cite{sonka2014image}. However, such devices often exhibit reduced precision outdoors due to sunlight interference, as sunlight has higher power compared to the emitted signal~\cite{hansard2012time, tolgyessy2021evaluation}.

\subsection{Sources of Noise and Errors in 3D Scanning}
Scanning devices in real life are prone to various sources of noise and errors, related to the environmental conditions and limitations of underlying technology.

Temporal noise in 3D scanning devices refers to variations in the captured data over time, introducing fluctuations or inconsistencies in the measurements. Temporal noise is often correlated between consecutive scans and can arise from a range of factors, including electronic instability, sensor characteristics, or environmental conditions~\cite{mallick2014characterizations}. The amount of temporal noise can also be influenced by colour and material properties of the observed objects~\cite{falie2007noise, tolgyessy2021evaluation, vogt2021comparison, wasenmuller2017comparison} and the geometry of the scene~\cite{mallick2014characterizations}.

The presence of a different source of similar radiation can interfere with the device's ability to correctly calculate the distance of the objects in the scene. Such interference can be caused by ambient light~\cite{el2012study}, radiation from other active imaging devices~\cite{berger2011markerless, butler2012shake} or even radiation emitted from the device itself when the scene contains reflective surfaces~\cite{sarbolandi2015kinect}.

%Mitigating temporal noise is crucial for achieving high-quality and consistent 3D scans, often requiring advanced signal processing techniques, calibration procedures, or environmental controls to enhance the overall reliability and precision of the scanning device.

Systematic errors may also arise during 3D scanning. This type of errors result in consistent differences between the scans and the actual scene geometry. For SL cameras, it is mainly caused by inadequate calibration, low resolution, and coarse value quantisation~\cite{khoshelham2012accuracy}. In the case of ToF cameras, the measurement is based on mixing of different optical signals and approximation of their shapes. The mentioned approximation is one of the contributions to the effect referred to as wiggling~\cite{sarbolandi2015kinect}, periodic change of the systematic error with distance. Both SL and ToF cameras may also suffer from temperature drift~\cite{tolgyessy2021evaluation, wasenmuller2017comparison}. Systematic error of devices can be modelled well when precise information about the scene is known~\cite{tolgyessy2021evaluation}.

\subsection{Training NNs using Synthetic Data}

In the context of machine learning and neural network training, the fusion of synthetic data generation, domain randomization and data augmentation can be leveraged as powerful tools to avoid expensive creation of real datasets. 

A widely recognized tool for generating synthetic data is for example {NVIDIA} replicator\footnote{\url{https://developer.nvidia.com/omniverse/replicator}}. Synthetic data can further be enhanced by GANs~\cite{Duplevska2022GANSynthetic}, analytical emulation of known imaging errors, artifacts and noise~\cite{kocur2023correction}, or domain randomization which introduces variability by altering key factors such as object properties, lighting conditions, and camera perspectives~\cite{tremblay2018training}.

%Furthermore, synthetic and real data can be further used through for next generative pipeline and domain randomization. Through synthetic data generation with GANs~\cite{Duplevska2022GANSynthetic}, researchers can create artificial datasets that closely mimic real-world scenarios, facilitating the training of neural networks in environments where obtaining extensive real data is impractical. Domain randomization introduces variability by altering key factors such as object properties, lighting conditions, and camera perspectives, ensuring that the model becomes adept at handling diverse and unpredictable situations~\cite{tremblay2018training}.

\section{Estimating 3D Camera Noise Parameters}

In this section we describe the process of estimating the parameters of two types of noise occurring in real 3D scans and their dependence on the distances of objects as well as the angle of the imaged surfaces. In \autoref{sec:synth_experiment} we perform an experiment showing that the estimated parameters can be used to improve the performance of models trained on synthetic data.

\subsection{Lateral and Axial Noise}

We specifically investigate two types of noise: lateral and axial. These are the two most dominant types of noise present in real 3D scans.

\begin{figure}
    \centering
    \begin{subfigure}{.49\columnwidth}
        \centering
        \includegraphics[height=3cm]{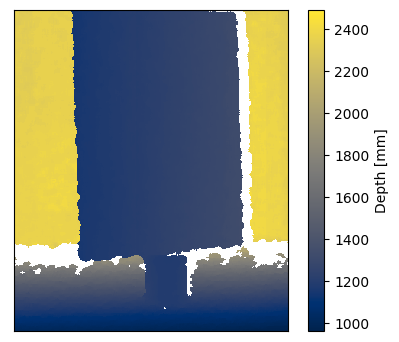}
        \caption{}
        \label{fig:lateralExample}
    \end{subfigure}
    \begin{subfigure}{.49\columnwidth}
        \centering
        \includegraphics[height=3cm]{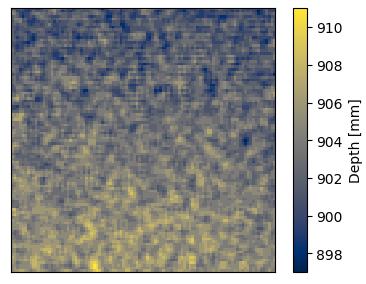}
        \caption{}
        \label{fig:axialExample}
    \end{subfigure}
    \caption{(a) Cropped range image of a white paper (blue rectangular area) positioned 1.25 m away from the camera at a 20° angle captured by Kinect v1. White pixels represent missing values. Lateral noise can be seen at the paper boundaries which are straight in the real scene.
    (b) Cropped range image of a planar wall captured by Kinect v2 at 90 cm distance with notable axial noise.}    
    \vspace{-1.5ex}
\end{figure}

%\begin{figure}
%\centerline{\includegraphics[width=0.8\columnwidth]{images/lateral example}}
%\caption[]{Cropped range image of a white paper (blue rectangular area) positioned 1.25 m away from the camera at 20° angle captured by Kinect v1. White pixels represent missing values. Lateral noise can be seen at the paper boundaries which are straight in the real scene.}
%\label{fig:lateralExample}
%\end{figure}

%\begin{figure}
%\centerline{\includegraphics[width=0.8\columnwidth]{images/axial example}}
%\caption[]{Cropped range image of a planar wall captured by Kinect V2
%\nameref{para:KinV2} 
%at 90 cm distance.}
%\label{fig:axialExample}
%\end{figure}

\paragraph{Lateral noise}
\label{sec:Alateral}
Lateral noise refers to error in the reported position in the camera's $xy$-plane. Even though lateral noise affects all measurements, it is most visible at object boundaries, as illustrated in \autoref{fig:lateralExample}. 
%The pictured range image contains a vertically positioned paper. Although the paper has perfectly straight edges, the edges of the paper in the range image are not straight. Similarly, the edge of the shadow (missing values indicated by white pixels) to the right of the paper is also irregular. 
Existing research~\cite{nguyen2012modeling, mallick2014characterizations} suggests the distance of the object and its angle influences the amount of lateral noise. 

\paragraph{Axial noise}
\label{sec:Aaxial}
Axial noise refers to noise orthogonal to the imaging plane, parallel to the $z$-axis of the camera. The lateral noise presents itself by altering the positions of depth values in the range image, while the axial noise can be observed in the individual depth values themselves. An example of axial noise is presented in \autoref{fig:axialExample}. 
%The depth measurements in the captured range image vary, while in reality, the wall was not as uneven.

Multiple factors are known to influence axial noise, from geometry of the scene to properties of the material of the surfaces in the scene~\cite{mallick2014characterizations, nguyen2012modeling, tolgyessy2021evaluation}. 

For SL cameras, according to pin-hole camera model and the disparity-depth model, the standard deviation of axial noise $\sigma_z$ increases quadratically with increasing depth and can be calculated as~\cite{mallick2014characterizations}:
\begin{equation}
\sigma_z = \left(\frac{m}{fb}\right) z^2 \sigma_\rho~,
\end{equation}
where $z$ refers to depth, $\sigma_\rho$ to the standard deviation of normalised disparity values, $f$ to the focal length, $b$ to the length of the baseline, and $m$ to the parameter of internal disparity normalisation. In this paper we estimate the noise levels for both axial and lateral noise directly from the observed data without relying on knowledge of the camera intrinsics.

%\begin{figure}
%\centerline{\includegraphics[width=0.3\textwidth]{images/cam plane small}}
%\caption[]{Illustration of camera space.}
%\label{fig:cameraPlane}
%\end{figure}

\subsection{Custom Dataset}

% \begin{figure}
%     \centering
%     \begin{subfigure}{.3\columnwidth}
%         \centering
%         \includegraphics[width=\linewidth]{images/KinectForWindows}
%         \caption{\nameref{para:KinV1}}
%         \label{fig:camsK1}
%     \end{subfigure}
%     \begin{subfigure}{.3\columnwidth}
%         \centering
%         \includegraphics[width=\linewidth]{images/Xbox-One-Kinect}
%         \caption{\nameref{para:KinV2}}
%         \label{fig:camsK2}
%     \end{subfigure}
%     \begin{subfigure}{.36\columnwidth}
%         \centering
%         \includegraphics[width=\linewidth]{images/MotionCam}
%         \caption{\nameref{para:PhoXi}}
%         \label{fig:camsMC}
%     \end{subfigure}
%     \caption{3D cameras.}
%     \label{fig:cams}
% \end{figure}

In order to estimate the levels of lateral and axial noise in various 3D scanning devices we collected a custom dataset. The dataset consists of scenes with a large planar surface (white rectangular cardboard) under various rotations.

We captured the scene using three 3D cameras:
\begin{itemize}
\setlength{\itemsep}{0.1em}
    \item \textbf{Kinect v1}\label{para:KinV1} utilises IR SL projector combined with a monochrome CMOS sensor for depth sensing, supplying range images with $640 \times 480$ resolution at 30 fps. Its default depth range is 0.8 m - 4.0 m, 0.4 m - 3.0 m in \textit{near mode}.
% The device is pictured in \autoref{fig:camsK1}. 

    \item \textbf{Kinect v2}\label{para:KinV2} employs a ToF camera for depth sensing. Compared to its predecessor, it has a wider field of view and offers depth measurements with greater accuracy and wider depth range, 0.5 m - 4.5 m. The resolution of the range images is, however, slightly smaller, $512 \times 424$. %The device can be viewed in \autoref{fig:camsK2}.

    \item \textbf{MotionCam-3D}\label{para:PhoXi} camera by Photoneo is based on SL range sensing. Thanks to parallel structured light technology
    %\footnote{\url{https://www.photoneo.com/the-revolution-in-machine-vision/}}
    ~\cite{photoNeoPattent}, the camera is able to capture dynamic scenes. Overall, the camera offers resolution up to $1680 \times 1200$. The MotionCam-3D can run in two different modes, the static scanner mode where the resolution and scanning time are higher, and dynamic camera mode where the scanning time and the output resolution are lower. 
 
\end{itemize}

\begin{figure}
    \centering
    \begin{subfigure}{.16\columnwidth}
        \centering
        \includegraphics[height=3cm]{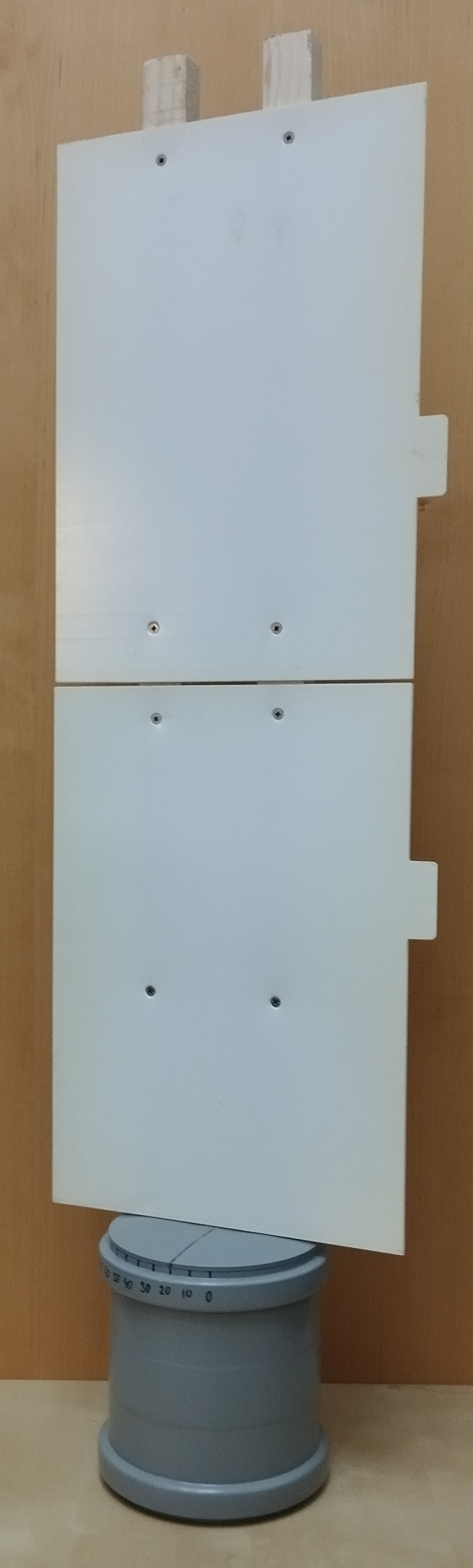}
        \caption{stand}
        \label{fig:stand}
    \end{subfigure}
    \begin{subfigure}{.43\columnwidth}
        \centering
        \includegraphics[height=3cm]{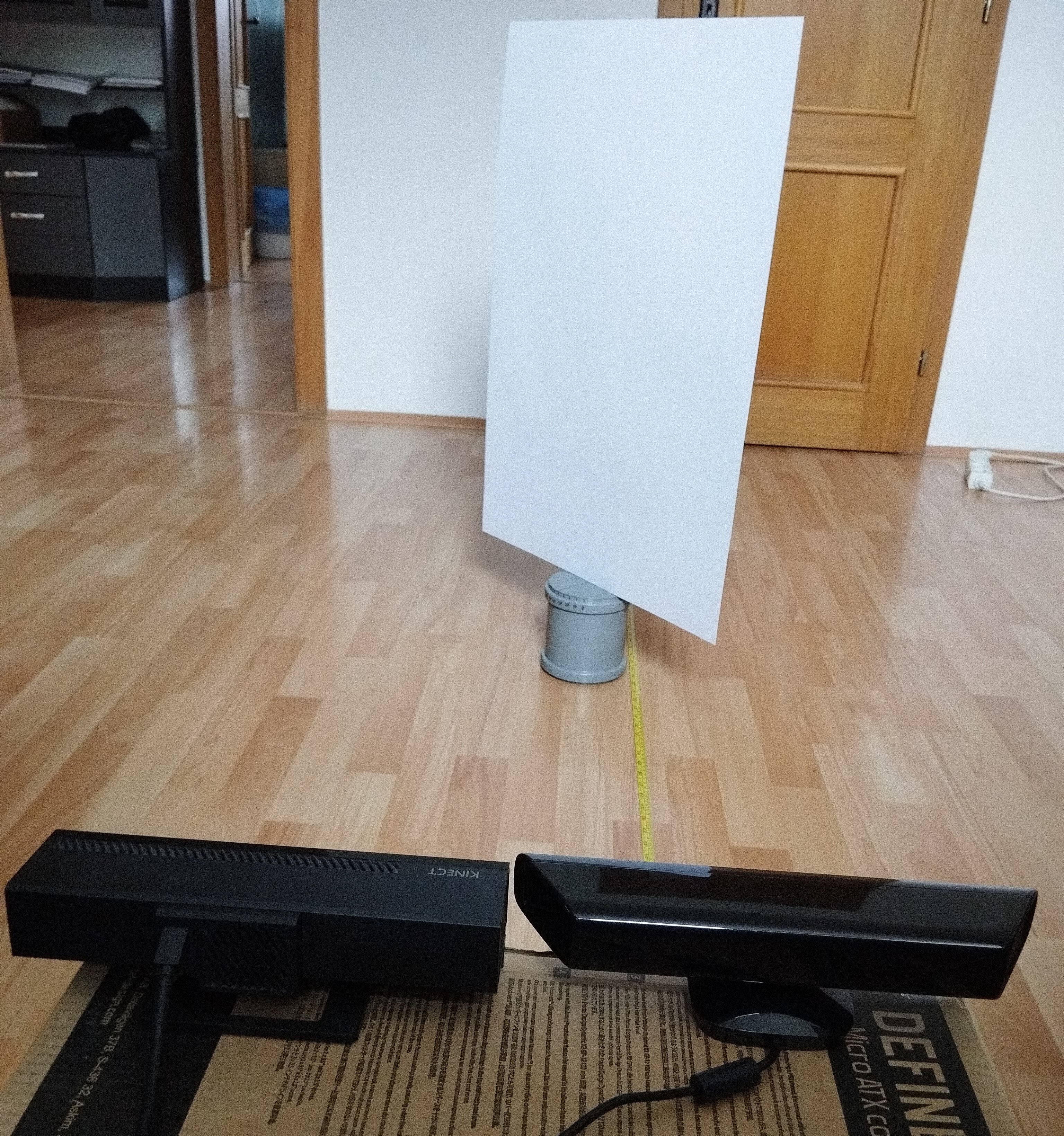}
        \caption{Kinect v1 and v2}
        \label{fig:anglesSetupK}
    \end{subfigure}
    \begin{subfigure}{.36\columnwidth}
        \centering
        \includegraphics[height=3cm]{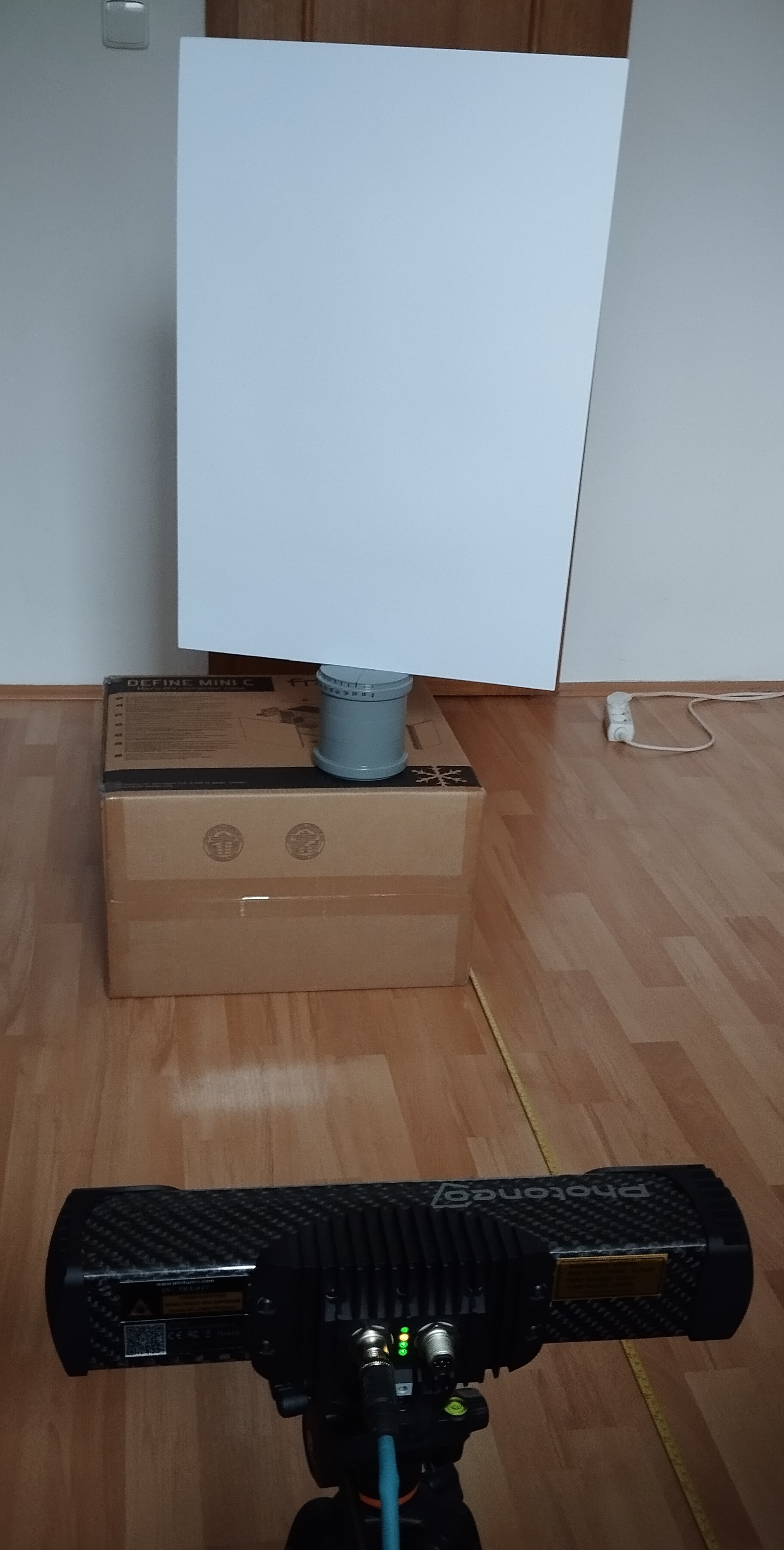}
        \caption{MotionCam-3D}
        \label{fig:anglesSetupMC}
    \end{subfigure}
    \caption{Physical setup for capturing surface at different angles and distances.}
    \label{fig:anglesSetup}
    \vspace{-1.5ex}    
\end{figure}

\begin{figure}
    \centering
    \begin{subfigure}{.28\columnwidth}
        \centering
        \includegraphics[height=18mm]{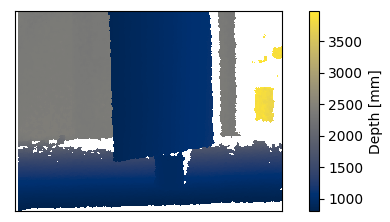}
        \caption{Kinect v1}
        \label{fig:imgDataKin1}
    \end{subfigure}
    \begin{subfigure}{.28\columnwidth}
        \centering
        \includegraphics[height=18mm]{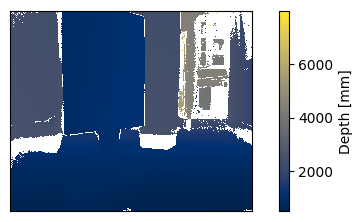}
        \caption{Kinect v2}
        \label{fig:imgDataKin2}
    \end{subfigure}
    \begin{subfigure}{.35\columnwidth}
        \centering
        \includegraphics[height=18mm]{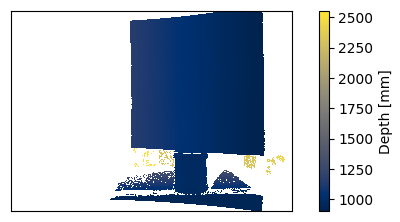}
        \caption{MotionCam-3D}
        \label{fig:imgDataMC}
    \end{subfigure}
    \caption{Range images captured by devices. The scene contains a white paper at 1 m distance and 30° angle captured as portrayed in \autoref{fig:anglesSetup}.}
    \label{fig:examples}
    \vspace{-1.5ex}
\end{figure}

To mitigate the effects of thermal drift the devices were warmed up by capturing range images of a blank wall in 1-minute intervals for 60 minutes prior to collecting the samples in the dataset. 

To investigate the influence of surface distance and angle on noise a set of range images containing a white planar paper at various positions was captured. To minimise any distortion of the paper, heavy-weight card stock was mounted on a rigid stand, displayed in \autoref{fig:stand}. The stand is comprised of two plastic boards mounted to two wooden beams attached to a wide plastic pipe with a plug. The rubber seal between the pipe and the plug was shaved to allow smooth rotation while preserving the position when idle. The stand was constructed to have the centre of rotation in the horizontal centre of the paper, with markings noting the rotation angle. 

With a mounted paper, this stand was positioned at various distances from the cameras and was rotated for the capture of various scenes. For each such stationary scene 200 range images were captured by each camera. 
To minimise the impact of temporal noise, for each set of range images capturing one scene, an average range image was computed by averaging captured depth values for individual pixels. To ensure all the cameras captured the same scene, the entire process was repeated, as all the cameras did not reasonably fit into the same space at once. The setups are portrayed in \autoref{fig:anglesSetup}, while examples of captured range images are displayed in \autoref{fig:examples}.

% Additionally, 3000 range images capturing a blank wall and 500 range images containing black and rainbow papers each were also captured.

In order to estimate the effects of angle and distance of planar surfaces on noise levels we segment the paper in the range images using manual annotation in conjunction with the Canny edge detection~\cite{canny1986computational} and Hough transformation~\cite{duda1972use}.

\subsection{Lateral Noise Estimation}

\begin{figure}
\centerline{\includegraphics[width=0.9\columnwidth]{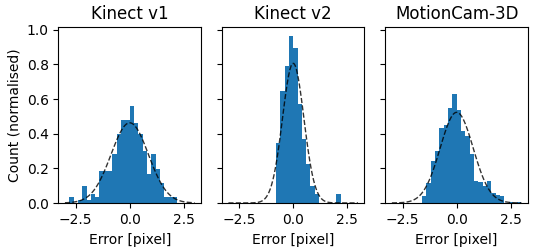}}
\caption[]{Normalised histograms of lateral error values. Collected from 200 images by Kinect v1 and Kinect v2, and 100 images by MotionCam-3D. Each histogram represents a scene containing the white paper at 0° angle. The distances differ for each camera, being the shortest at which the paper was captured completely; 1m for Kinect v1, 0.75 m for Kinect v2, 0.5 m for MotionCam-3D. Each histogram contains fitted normal distribution (dashed line).}
\label{fig:lateralHists}
    \vspace{-1.5ex}
\end{figure}

\begin{figure}
\centerline{\includegraphics[width=\columnwidth]{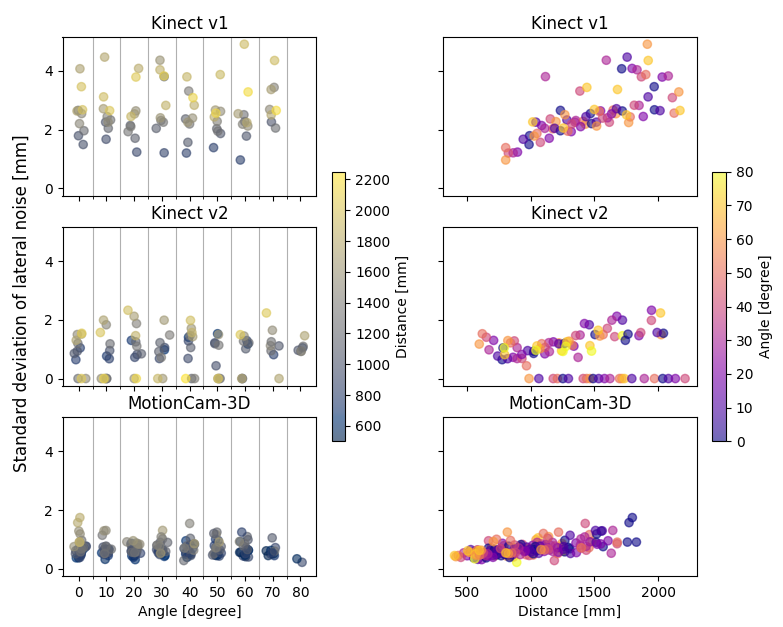}}
\vspace{-1.5ex}
\caption[]{Visualisation of the relationship between the standard deviation of lateral noise, measured in mm, surface angle (left column), and distance (right column). Each row contains data from a different device. The plots in each column and row share the x and y axes respectively. In the plots of the left column, the underlying angle values are all multiples of 10. A random shift of horizontal position between frames was added for legibility.}
\label{fig:lateralGridmm}
\vspace{-1.5ex}
\end{figure}

To estimate the lateral noise levels we focus on the paper boundary. We first estimate the position of the boundary by fitting a line using orthogonal distance regression on the edge pixels. We perform this regression jointly for all scans with a given scene setup. We then calculate the distances of the edge pixels from the estimated boundary line.

Example histograms of the distances from the line fit are shown in \autoref{fig:lateralHists}. The Kolmogorov-Smirnov test rejected the normality of the distribution, probably due to the effects of quantization in pixel positions. However, we note that the error distributions closely resemble normal distributions.

As seen in \autoref{fig:lateralGridmm}, the level of lateral noise does not significantly change with surface angle. Previous research indicates hyperbolic increase of lateral noise at angles greater than 60° for Kinect v1~\cite{nguyen2012modeling}. Our experiments did not indicate such increase, however, thanks to large number of invalid pixels, we were not able to capture data for angles greater than 70°, and subsequently extract lateral noise. MotionCam-3D exhibited similar inability to capture surfaces at extreme angles. Contrastingly, Kinect v2 had no problem with 80° angle and exhibited no increase in noise with increasing angle. 

A slight decline in the standard deviation with rising angle can be observed. We note that this decline may not be caused by the change in angle directly, but as a result of presence of other noise causing a great number of invalid pixels and thus preventing lateral noise analysis. This type of noise increases by rising distance, as surfaces with progressively lower angles with the camera view are affected. Hence, surfaces at greater angles are harder to measure from greater distances, leaving fewer samples resulting in lower standard deviation.

Unlike in the case of the paper's angle, the standard deviation of the errors is not constant throughout all distances, as seen in \autoref{fig:lateralGridmm}. Noteworthy is the elevated standard deviation at shorter distances, between 50 cm and 1 m, for Kinect v2 and MotionCam-3D. Kinect v1 was not able to capture the paper at such short distances at all. The standard deviation of errors in millimetres increases linearly with increasing distance, at different rate for each camera, depending on the camera's physical parameters~\cite{nguyen2012modeling}. Note that this is equivalent to the standard deviation remaining constant under varying distances when measured in pixel coordinates.

By aiming to capture the scenes simultaneously with multiple cameras, the position of the paper was not always perfectly centred for all cameras. As a result, for the Kinect v2, the paper's right edge was much closer to the centre than the left edge. On multiple occasions, the right edge was captured as a perfectly vertical line in all 200 images captured for the scene, while the left edge was not. 
%An example of such instance can be viewed in \autoref{fig:lateralPerfectRightEdge}. This resulted in the standard deviation of the error being equal to zero. The values are present in the middle row plots of \autoref{fig:lateralGrid} and \autoref{fig:lateralGridmm}. All of these measurements correspond to the right edge. 
This can be observed in \autoref{fig:lateralGridmm} as some values are reported with standard deviation of 0. From our limited data, a correlation of lateral noise with the pixel's position seems likely. Further experimentation would be required to fully explore this relationship. 

The results show that MotionCam-3D exhibits overall lower levels of lateral noise than both Kinect cameras with Kinect v2 achieving lower noise levels of the two.

%The aim of this study is to verify whether adding noise based on an estimated noise model to synthetic training data in a deep learning pipeline is beneficial for outcomes on real data. To obtain such model we fit polynomials of degree two on the data presented in \autoref{fig:lateralGridmm}. As our model does not take pixel position into consideration, the data points resulting zero standard deviations for were omitted in fitting a model of the lateral noise for Kinect v2. 

% \begin{figure}
% \centerline{\includegraphics[width=0.6\columnwidth]{images/lateral perfect with centre}}
% \caption[]{Cropped range image of a white paper positioned 1.5m away from the camera at 0° angle captured by \nameref{para:KinV2}. Horizontal centre of the whole range image is marked by the vertical red line.}
% \label{fig:lateralPerfectRightEdge}
% \end{figure}

\subsection{Axial Noise Estimation}

\begin{figure}
\centerline{\includegraphics[width=\columnwidth]{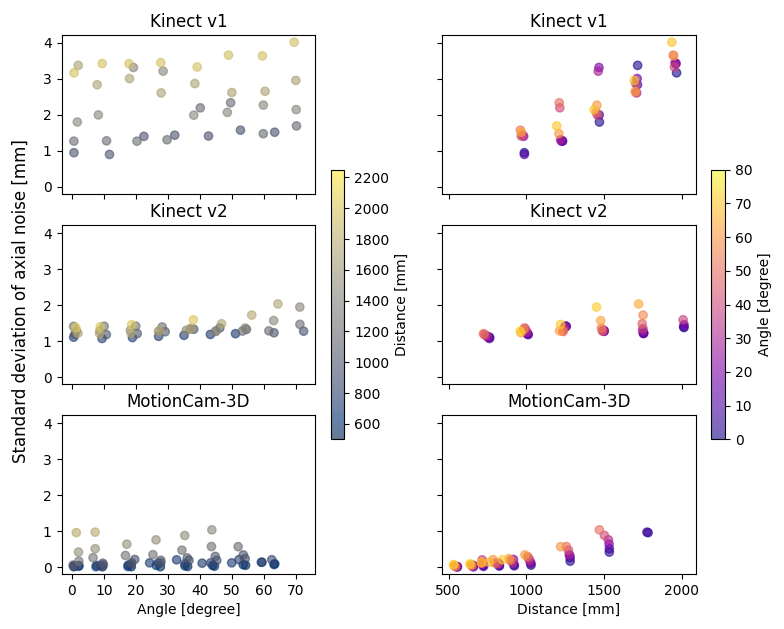}}
\vspace{-1.5ex}
\caption[]{Visualisation of the relationship between standard deviation of axial noise, the surface angle (left column), and the distance (right column). Each row contains data from a different device. The plots in each column and row share x and y axes.}
\label{fig:axialFourier}
\vspace{-1.5ex}
\end{figure}

To obtain the distributions of axial noise we first performed low-pass filtering jointly on all scans of scenes with the same scanner distances and angles. We then calculated the standard deviations of differences of depth from the values obtained by filtering. We have opted for this approach as despite using heavy stock paper, the paper surface was not perfectly planar. We have also tested different types of filtering which led to similar results.

Similar to lateral noise, the relationship between angle and distance on the standard deviation of noise has been investigated. 
The results are visualised in \autoref{fig:axialFourier}. MotionCam-3D exhibits least axial noise, followed by Kinect v2 and Kinect v1 with the greatest magnitude of noise.

From the right column in \autoref{fig:axialFourier}, the influence of surface distance on the standard deviation can be clearly seen for both SL cameras, Kinect v1 and MotionCam-3D. For Kinect v2 camera, the standard deviation does not change much with increasing distance compared to the other two cameras. The influence of surface angle can also be seen. In the case of Kinect v1, the values of standard deviation seem to fluctuate unpredictably with changing angle. This may be caused by different sources of noise such as systematic noise arising from the imaging process.
%This is caused by the pattern in the noise described in Section \ref{sec:temporal}, which affected different portions of the observed surface as the angle changed.

\subsection{Noise Models}

\label{sec:noise_models}

\begin{figure}
\centerline{\includegraphics[width=\columnwidth]{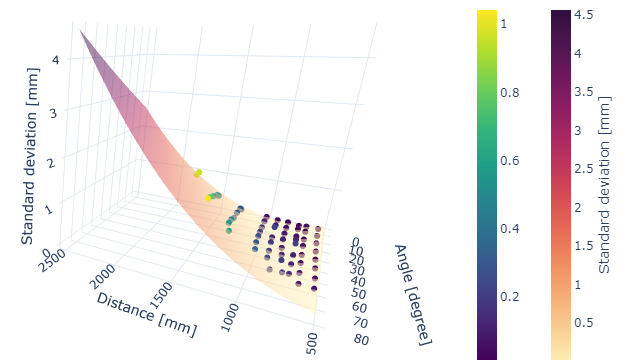}}
\caption[]{Fitted polynomial function of degree 2 for axial noise of MotionCam-3D, displayed as the surface, with the measured values, displayed as points colored by respective standard deviation of the sample.}
\label{fig:axialFit}
\vspace{-1.5ex}
\end{figure}

In previous subsections we have shown that standard deviations of both types of noise depend on both the distance of objects to the scanners as well as the angle of the imaged surface. To model the noise we fit the data shown in \autoref{fig:lateralGridmm} and \autoref{fig:axialFourier} with degree two polynomials using the ordinary least squares method. The resulting coefficients for both lateral $\sigma_L$ and axial noise $\sigma_z$ are in \autoref{tab:noises}. The resulting fit for the axial noise of MotionCam-3D is shown in \autoref{fig:axialFit}.
. 

\begin{table*}
    \centering
    \caption{Fitted standard deviations of lateral noise ($\sigma_L$ - in pixels) and axial noise ($\sigma_z$ - in millimeters). Parameter $\theta$ represents the surface angle and $z$ the distance from the camera center.}
    \vspace{-1.5ex}
    \label{tab:noises}
        \begin{tabularx}{\textwidth}{ l||m{2.3cm} X }
            \hline
            Kinect v1 & \small$\sigma_L\left(z, \theta\right)[px]=$ & \tiny$
              0.94 
            + 4.51 \cdot 10^{-5} \cdot z 
            + 6.20 \cdot 10^{-4} \cdot \theta$\\
            \hline
            Kinect v2 & \small$\sigma_L\left(z, \theta\right)[px]=$ & \tiny$
              0.736 
            - 6.20 \cdot 10^{-4} \cdot z 
            + 5.35 \cdot 10^{-3} \cdot \theta 
            + 2.13 \cdot 10^{-7} \cdot z^2 
            - 1.40 \cdot 10^{-6} \cdot z \cdot \theta 
            - 4.13 \cdot 10^{-5} \cdot \theta^2$\\
            \hline
            MotionCam-3D & \small$\sigma_L\left(z, \theta\right)[px]=$ & \tiny$
              0.915 
            - 6.91 \cdot 10^{-5} \cdot z 
            + 2.84 \cdot 10^{-3} \cdot \theta$\\
            \hline
            Kinect v1 & \small$\sigma_z\left(z, \theta\right)[mm]=$ & \tiny$
            - 0.422 
            + 6.89 \cdot 10^{-4} \cdot z 
            + 2.24 \cdot 10^{-2} \cdot \theta 
            + 5.99 \cdot 10^{-7} \cdot z^2 
            - 2.70 \cdot 10^{-6} \cdot z \cdot \theta 
            - 1.52 \cdot 10^{-4} \cdot \theta^2$\\
            \hline
            Kinect v2 & \small$\sigma_z\left(z, \theta\right)[mm]=$ & \tiny$
              1.17 
            + 9.72 \cdot 10^{-5} \cdot z 
            - 1.37 \cdot 10^{-2} \cdot \theta 
            - 6.35 \cdot 10^{-9} \cdot z^2 
            + 7.86 \cdot 10^{-6} \cdot z \cdot \theta 
            + 1.17 \cdot 10^{-4} \cdot \theta^2$\\
            \hline
            MotionCam-3D & \small$\sigma_z\left(z, \theta\right)[mm]=$ & \tiny$
              0.599
            - 1.43 \cdot 10^{-3} \cdot z 
            - 8.94 \cdot 10^{-3} \cdot \theta 
            + 8.84 \cdot 10^{-7} \cdot z^2 
            + 1.27 \cdot 10^{-5} \cdot z \cdot \theta 
            + 2.75 \cdot 10^{-5} \cdot \theta^2$\\
            \hline
        \end{tabularx}
\end{table*}

\section{Enhancement of Synthetic Training Data with Emulated Noise}

\label{sec:synth_experiment}

In this section we present an experiment that verifies the importance of selecting an optimal level of noise when generating synthetic training data for deep neural network training. We evaluate the effects of noise on a simple segmentation task. We train the networks on synthetic data and evaluate them on real-world scans.

\begin{figure*}
    \centering
    \begin{subfigure}{.24\linewidth}
        \centering
        \includegraphics[width=0.9\linewidth]{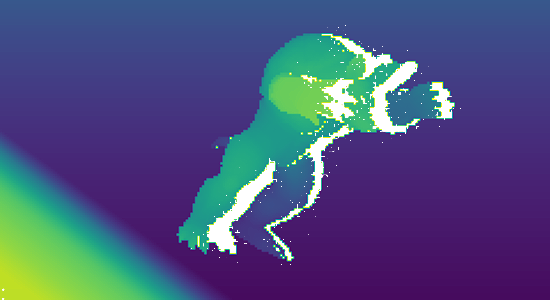}
        \caption{$M_n = 0$}
    \end{subfigure}
    \begin{subfigure}{.24\linewidth}
        \centering
        \includegraphics[width=0.9\linewidth]{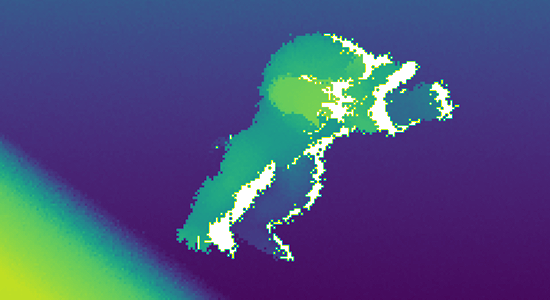}
        \caption{$M_n = 1$}
    \end{subfigure}
    \begin{subfigure}{.24\linewidth}
        \centering
        \includegraphics[width=0.9\linewidth]{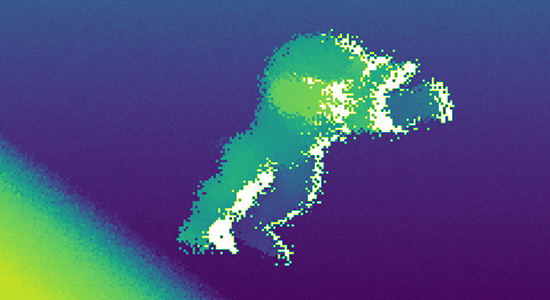}
        \caption{$M_n = 2$}
    \end{subfigure}
    \begin{subfigure}{.24\linewidth}
        \centering
        \includegraphics[width=0.9\linewidth]{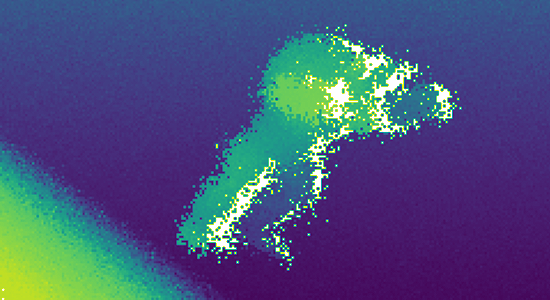}
        \caption{$M_n = 3$}
    \end{subfigure}
    \vspace{-1.5ex}
    \caption{Synthetic sample from our data with varying amount of lateral noise added to range images.}
    \label{fig:1}
    \vspace{-1.5ex}
\end{figure*}

\subsection{Real Evaluation Dataset}

To create our real data we manufactured five 3D models of the Stanford Armadillo. The objects were printed on J750 using the Vero family of materials. This allowed us to capture 55 real scans using 3 different variants of the MotionCam-3D. The real data contain samples from a close distance of around 70 cm, mid-range captures from around 100 cm, and longer-range shots from 150 cm. This should model the various use cases of the 3D scanning device, with varying amounts of noise. Apart from the Armadillos, various cuboid-shaped objects were included in the scene, some of which had a slightly reflective material causing further noise. The real data was split into a validation set with 20 samples, a test set of 25 captures, and 10 samples were used for training.

\subsection{Training Data}

To evaluate the benefit of adding axial and lateral noise into synthetic data, we have rendered training data for the task of object segmentation using specialized data generator~\cite{Gajdosech2022}, implemented to simulate the MotionCam-3D and other Photoneo scanners. We are dealing with a simplified setting - segmentation of a singular object, the Armadillo figurine.\footnote{\url{http://graphics.stanford.edu/data/3Dscanrep/}} Due to a current limitation of the renderer, we were unable to account for the angle of the surface, thus the amount of noise is only affected by distance. This simplification should not hinder the evaluation, as per our analysis the surface angle does not greatly affect the standard deviation of the noise, but the amount of missing samples instead. Some samples also contain cuboid-shaped walls of containers, which served as boundaries for the physical simulation of placing the Armadillos into the scene.

The dataset contains 180 synthetic samples. Additionally, we have included 10 real samples, which helped to avoid over fitting and permitted longer training. 
%With a size of 110 training samples, 
The dataset was designed to empirically evaluate the generalization of UNet-like CNN~\cite{Ronneberger2015}.
%trained over purely synthetic data. 
As different types of noise are abundant in the real samples, a network trained on clean rendered data is often unable to generalize. %To test the performance with various amounts of lateral and axial noise., we have defined a noise multiplicator. We multiply sigmas for both lateral and axial noise derived from our analysis. In other words, with multiplicator$\;=0.00$ the data is noiseless, and with multiplicator$\;=1.00$ it contains precisely the amount of noise arising from our noise analysis. The effect of different multiplicator values for the synthetic image of surface normals is shown in \autoref{fig:1}. 

\subsection{Training}

A 4-channel input image with surface normals and range image was used as an input to the U-Net shaped CNN. We have performed purely stochastic training with batch size $=1$, Adam optimizer with $10^{-4}$ initial learning rate, and binary cross-entropy as the loss function. 
%We trained the model for 48 epochs on the training data. For evaluation on the test set, we selected the model which performed best on the real validation set.
The number of epochs was determined by a training callback. It observed the IoU on the real validation set and picked the best model, which was then evaluated on the test set.
%We trained the model for up to 48 epochs. The model with maximum IoU on validation set was picked using a training callback, which was then evaluated on test set. This was achieved using a training callback, saving the model with maximum IoU on validation set.

\subsection{Varying Noise Levels}

\begin{figure*}[t]
\centerline{\includegraphics[width=0.99\linewidth]{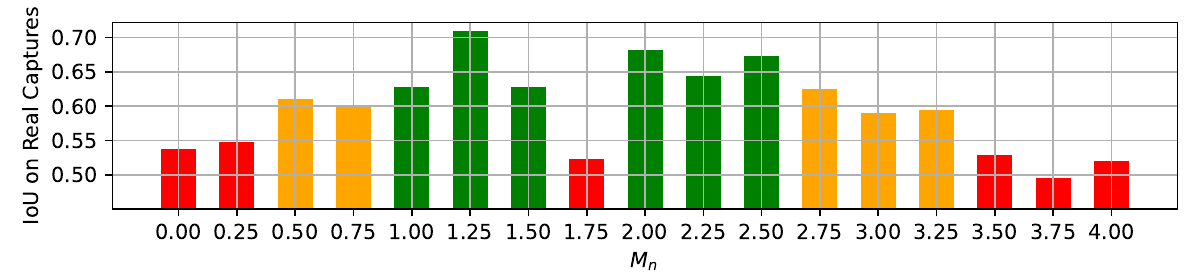}}
\vspace{-2.0ex}
\caption[]{Performance of neural network for object segmentation trained over data with varying amounts of noise. Multiplier $M_n$ of the sigma from our analysis is shown on the horizontal axis with $0.25$ interval. The resulting average IoU on a test set of real captures is visualized by height of the bars. For clarity, the top $6$ values are depicted in green, $6$ worst are in red and the remaining $5$ middle results are in orange.}
\label{graph}
\end{figure*}

\begin{figure}
    \centering
    \begin{subfigure}{.32\linewidth}
        \centering
        \includegraphics[width=0.9\linewidth]{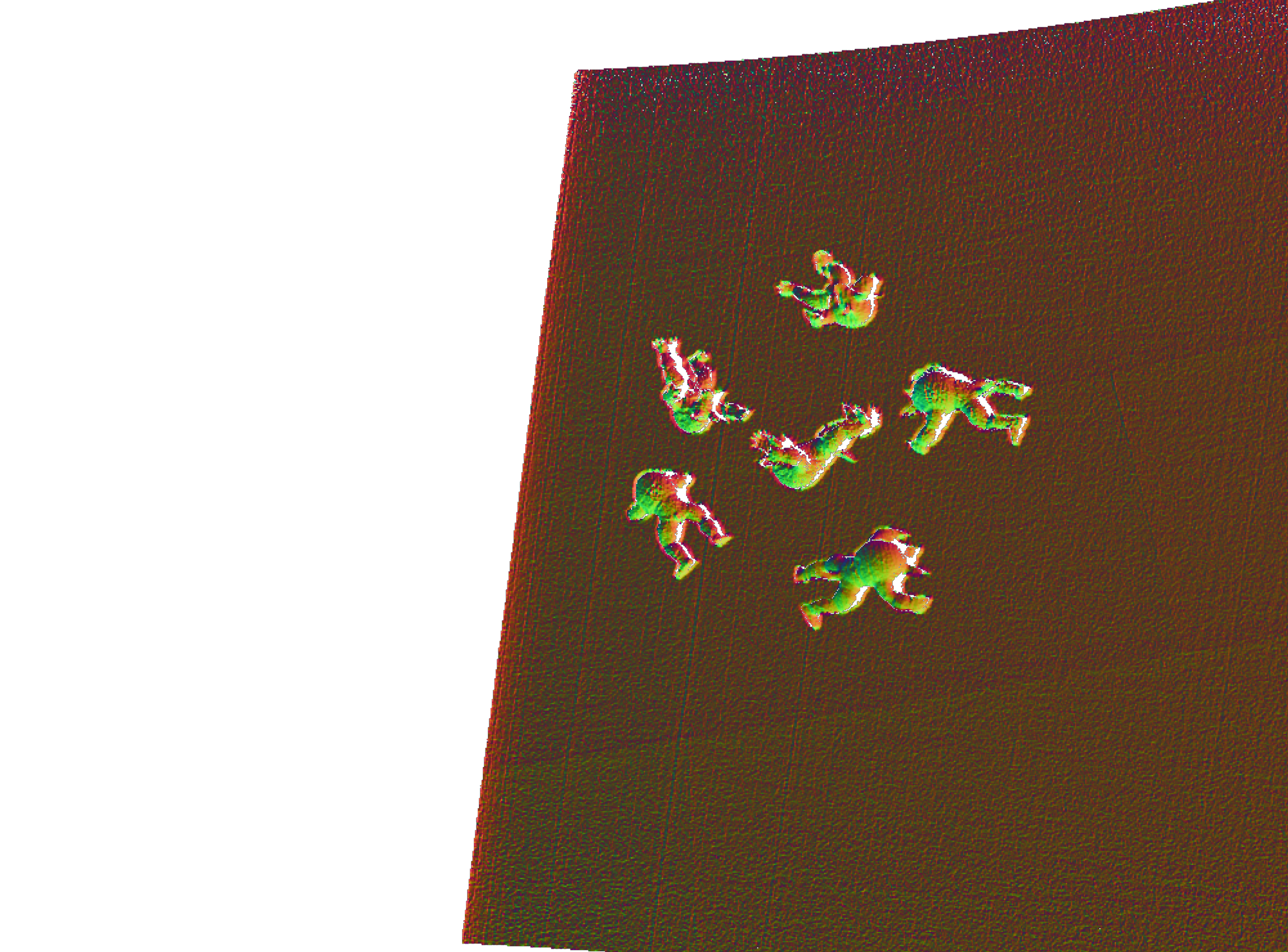}
        \caption{}
    \end{subfigure}
    \begin{subfigure}{.32\linewidth}
        \centering
        \includegraphics[width=0.9\linewidth]{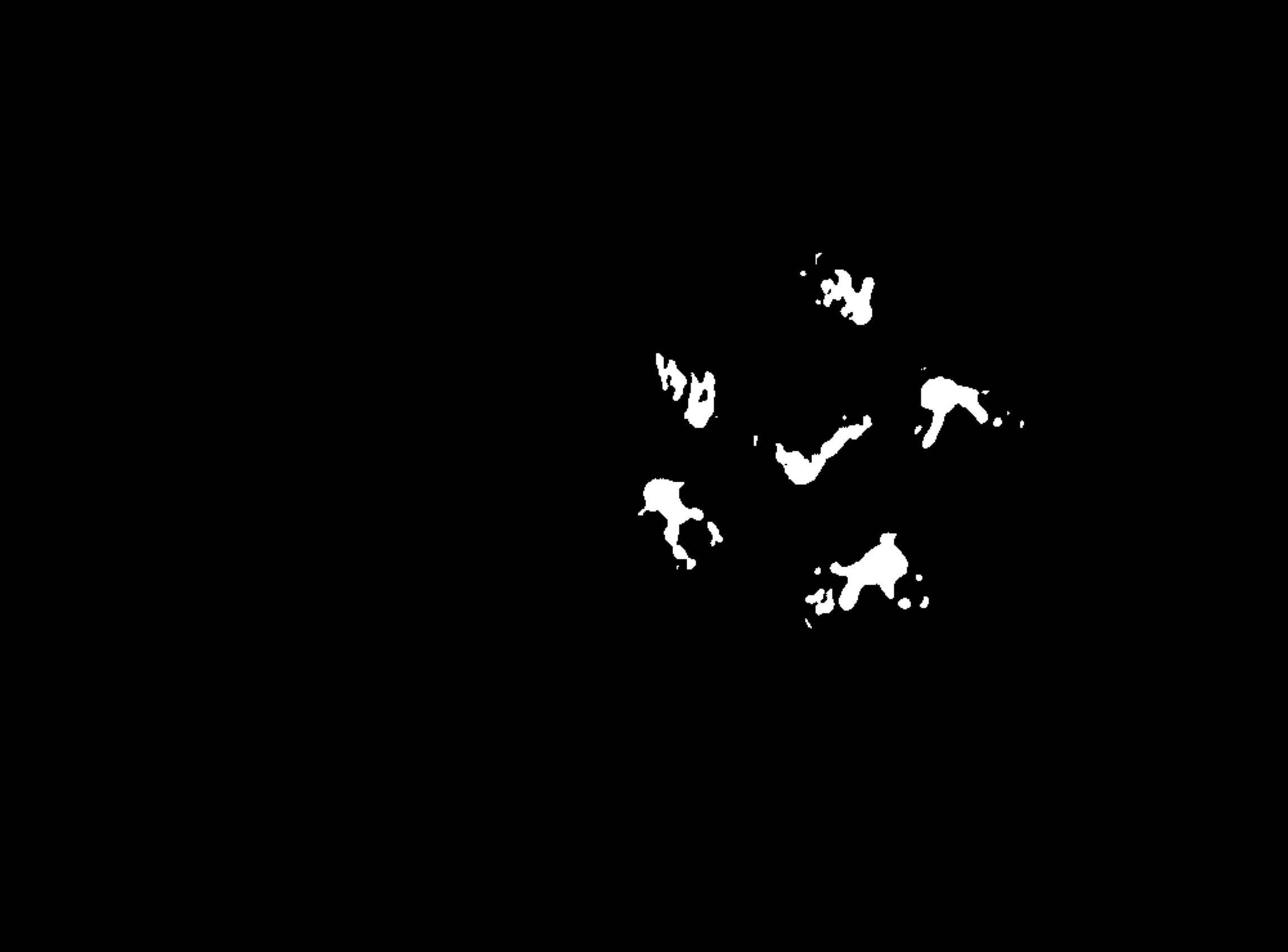}
        \caption{}
    \end{subfigure}
    \begin{subfigure}{.32\linewidth}
        \centering
        \includegraphics[width=0.9\linewidth]{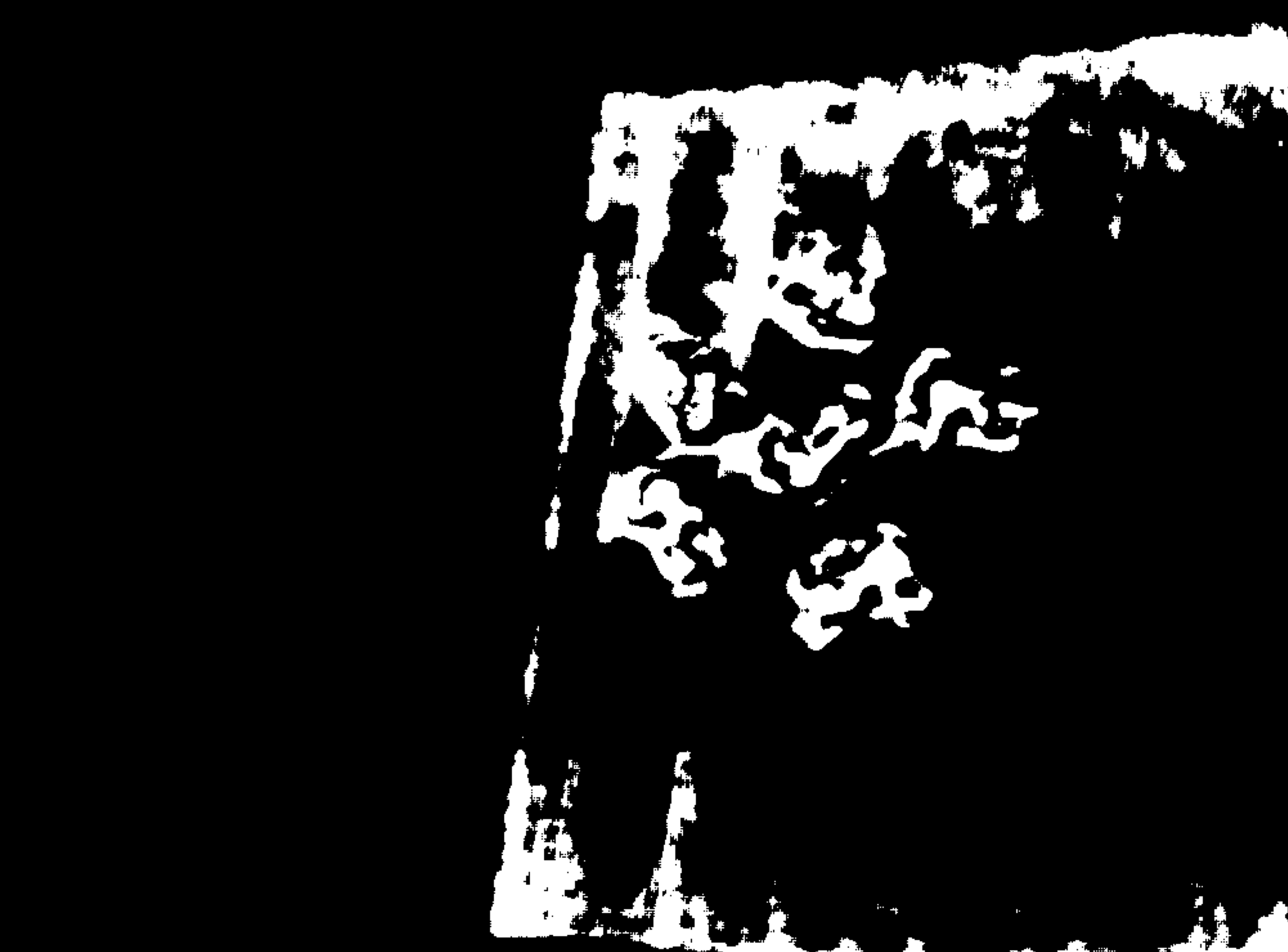}
        \caption{} 
    \end{subfigure}
    \vspace{-1.5ex}
    \caption{Qualitative evaluation on a real distant capture. As per our analysis, with larger distance, more noise is present, see normals in (a). In this situation, network trained on clean data without noise wrongly segments the rough surface as an object, see (c). On the other hand, network trained on data with noise ($M_n=1.25$) is resistant to the noise (b).}
    \label{fig:2}
    \vspace{-1.0ex}
\end{figure}

\begin{figure}
    \centering
    \begin{subfigure}{.32\linewidth}
        \centering
        \includegraphics[width=0.9\linewidth]{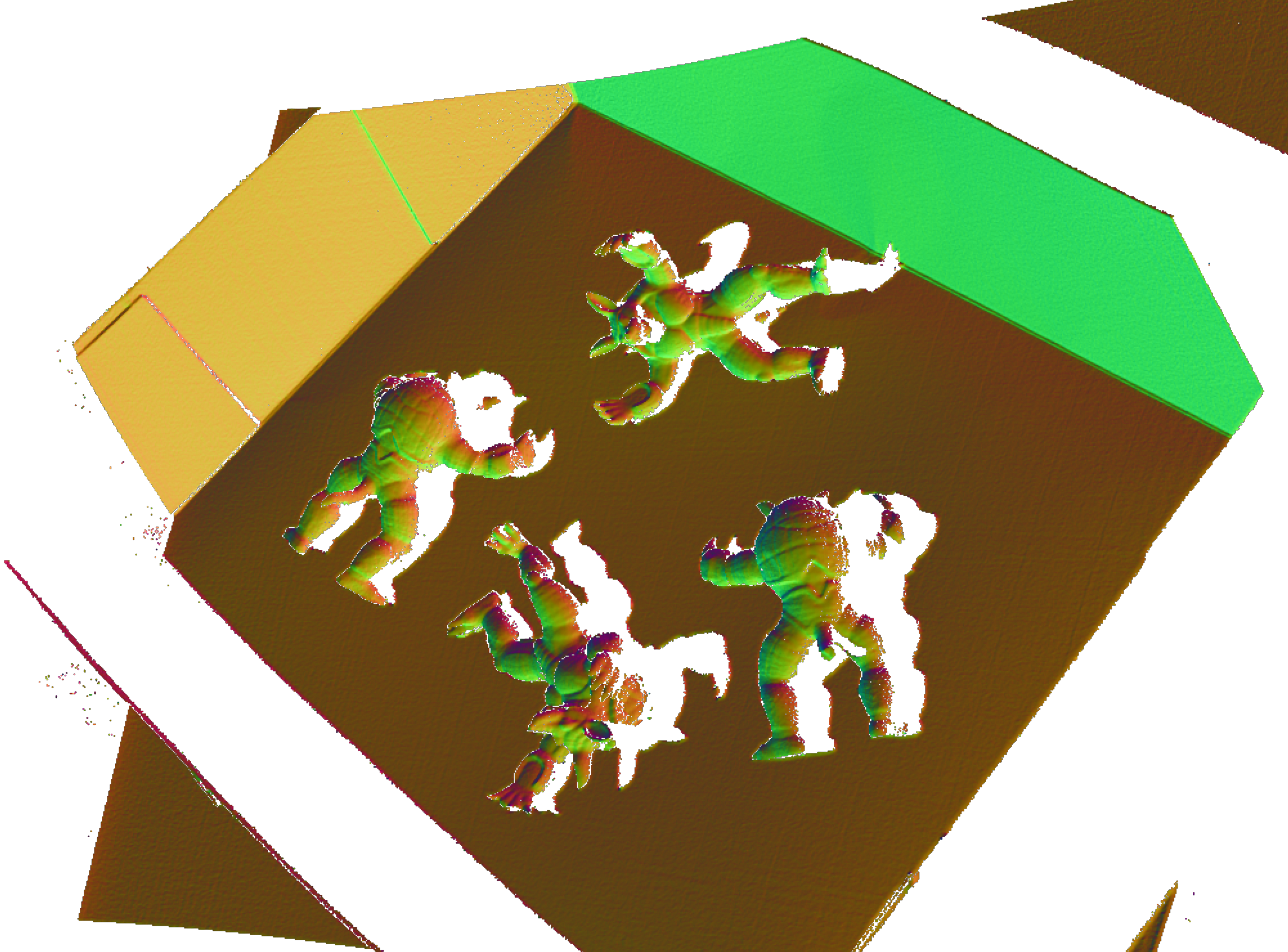}
        \caption{}
    \end{subfigure}
    \begin{subfigure}{.32\linewidth}
        \centering
        \includegraphics[width=0.9\linewidth]{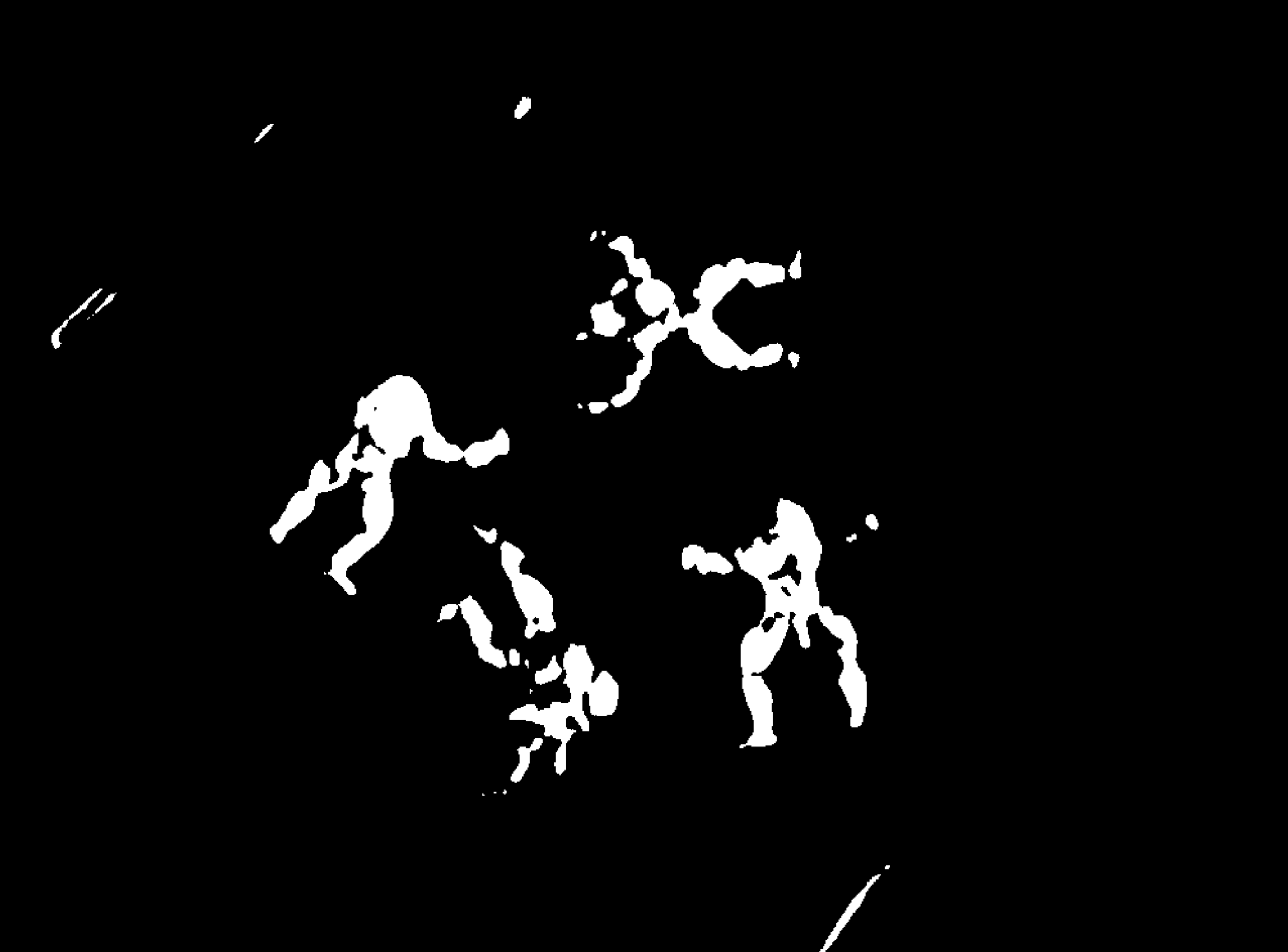}
        \caption{}
    \end{subfigure}
    \begin{subfigure}{.32\linewidth}
        \centering
        \includegraphics[width=0.9\linewidth]{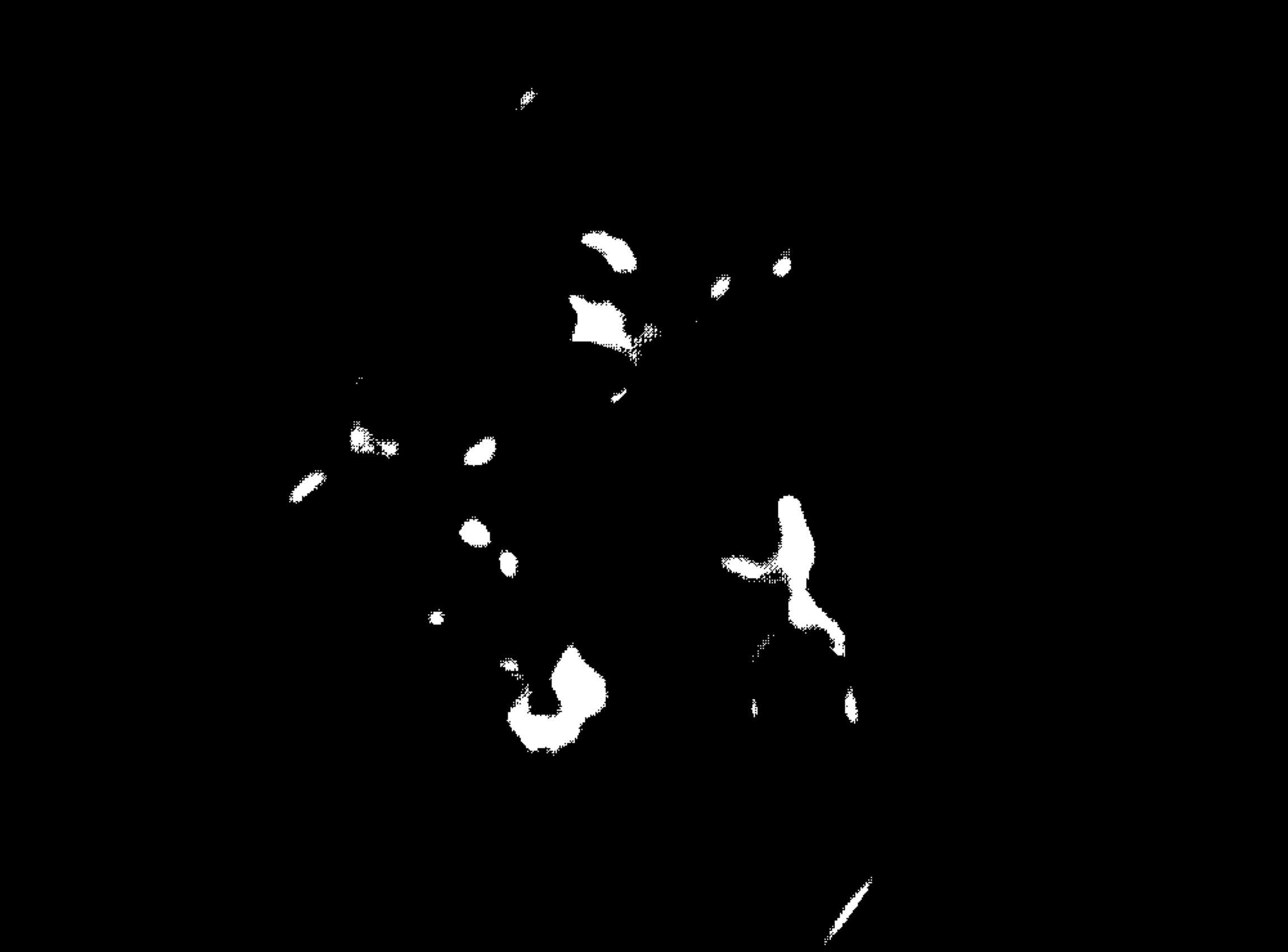}
        \caption{}
    \end{subfigure}
    \vspace{-1.5ex}
    \caption{Qualitative evaluation on a real scene captured in close distance. Image of surface normals used as input to the networks is shown in (a). The masks produced by the trained network are shown for (b) $M_n = 1.25$ (c) $M_n = 2$. Note the inability of the latter network to segment fine details.
    \vspace{-3mm}}
    \label{fig:3}    
    \vspace{-1.0ex}
\end{figure}

To verify the effect of various levels of emulated lateral and axial noise we train multiple models each using a different level of added noise. To control the noise we define a noise multiplicator:
\begin{equation}
M_n = \frac{\sigma_{synth}}{\sigma_{est}},
\end{equation}
where $\sigma_{est}$ denotes the estimated standard deviation of noise as presented in \autoref{sec:noise_models} and $\sigma_{synth}$ denotes the standard deviation of noise added to the training data. Note that the estimated standard deviations of noise depend on the surface angles and distances and the type of noise (axial, lateral), but the ratio $M_n$ is independent of these variables. The value of $M_n = 0$ indicates no noise added and $M_n = 1$ indicates noise added according to the levels estimated in the previous section. The effects of various values of $M_n$ on the produced synthetic samples are shown in \autoref{fig:1}.

%In this experiment, we have tested the combined effect of axial and lateral noise, with a rising amount by increasing our multiplicator in $0.25$ increment. See \autoref{graph} for the bar chart of the network's final performance over the test data. The network trained on synthetic data with slightly more noise (multiplicator$\;=1.25$) achieved the best result. Since we have modeled only 2 types of noises from the plethora of ones present in the real data, we hypothize that by the slight increase of the sigma arising from analysis, other noise types are implicitly modeled, effectively making the network more robust to noise uncaptured in the synthetic training data.\TODO{Viktor, tvoja myslienka, vieme tu nieco zacitovat?}

\subsection{Results}

We evaluated models for varying values of $M_n$ on the real testing data. \autoref{graph} shows the segmentation IoU metric for the evaluated models. The network trained on synthetic data with slightly more noise than estimated ($M_n=1.25$) achieved the best results. We hypothesize that by the slight increase of the $\sigma_{synth}$ arising from analysis, other noise types are implicitly modeled, effectively making the network more robust to noise uncaptured in the synthetic portion of the training data. This results indicates that adding slightly more noise than estimated allows the network to be more robust while retaining good accuracy.

The results also show an interesting second peak at $M_n=2$. By qualitative evaluation, we have verified that the network performance was better for scans captured from larger distances. In such cases, large amounts of interference noise is present, arising from the character of structured light technology and the presence of ambient lightning. This case is visualized in \autoref{fig:2}, where the network trained on data without noise wrongly segments the rough, noisy surfaces. 

On the other hand, we have samples captured from a close distance, where the captured surfaces are smoother and objects have sharper boundaries. In these situations, networks trained with greater noise levels ($M_n \ge 1.5$) have trouble with the segmentation of fine details and manage to only detect larger blobs of the objects, see \autoref{fig:3}. By a combined quantitative and qualitative analysis we conclude from our experiment that the network trained on data with noise $M_n=1.25$ delivers the most robust performance over various cases. Lastly, we note that setting $M_n = 1.75$ failed to both segment fine details in close-shots and was not as robust as networks trained for more extreme noise, delivering the weakest performance overall. 

Albeit limited in scope, the experiment presented in this section provides some insights into the effect of noise emulation during synthetic training data generation on real-world performance of the trained networks. Our experiment verifies the importance of noise inclusion in synthetic training data. Additionally, we can observe that adding too much noise may lead to poor models which are unable to detect fine details in the scene structure. We thus conclude that the ability to model noise as it occurs in real 3D cameras is an important aspect of synthetic training data generation. 

Our data\footnote{\url{https://doi.org/10.5281/zenodo.10581278}} and code\footnote{\url{https://doi.org/10.5281/zenodo.10581562}} used for noise analysis and network training is publicly available. 
%\footnote{\url{http://www.st.fmph.uniba.sk/~gajdosech2/cvww2024}}.
% do conclusion
%Potential future work includes the analysis of axial and lateral noise independently, inclusion of other noise models and verification over a larger dataset with multiple object classes. 

%To eliminate the need to overpower the amount of noise using an artificial multiplicator of sigma arising from analysis, other noise factors described in Section \ref{sec:related} must be modeled. Nevertheless, 

\section{Conclusion}

In this paper we have presented an approach for modeling axial and lateral noise of real 3D scanning devices. Using our proposed methodology it is possible to obtain a model of these types types of noise with respect to imaged object distance and surface angles. Knowledge of the noise parameters can be valuable when processing obtained 3D scans.

We also show that emulating noise when training a deep learning segmentation model on synthetic data is beneficial. Our experiment shows that the performance of the segmentation network on real data is best when the emulated noise is slightly stronger than estimated from the real scans. %The experiment additionally shows that adding too weak or excessive noise hurts performance.

%In future we intend to address the limitations of the presented work. 
In future, other types of noise should be modeled. Furthermore, the combined range image with surface normals should be compared to other data representations. We plan to expand the evaluation of the effects of noise levels with extended data as well as decoupling the effects of different types of emulated noise. Lastly, the interaction of added noise with other data augmentation techniques is worth investigating. 

\vspace{1em}
{\footnotesize \setstretch{1.0}
%\section*{Acknowledgements}
\noindent\textbf{Acknowledgments:}
The work presented in this paper was carried out in the framework of the TERAIS project, a Horizon-Widera-2021 program of the European Union under the Grant agreement number 101079338. Research result was obtained using the computational resources procured in the project National competence centre for high performance computing (project code:
311070AKF2) funded by European Regional Development Fund, EU Structural Funds Informatization of society, Operational Program Integrated Infrastructure. 
We thank Michal {Piovarči} for his help in preparing printing trays for our 3D models that were used for real dataset scanning.\par}

\newpage

{\small
\bibliographystyle{ieee}
\bibliography{literature}
}

%\begin{figure*}
%\centering
%\begin{subfigure}[t]{\dimexpr0.5\textwidth+20pt\relax}
%    \makebox[20pt]{\raisebox{72pt}{\rotatebox[origin=c]{90}{\nameref{para:KinV1}}}}
%    \includegraphics[width=\dimexpr\linewidth-20pt\relax]
%    {images/noisy KinectV1_crop}
%    \makebox[20pt]{\raisebox{72pt}{\rotatebox[origin=c]{90}{\nameref{para:KinV2}}}}
%    \includegraphics[width=\dimexpr\linewidth-20pt\relax]
%    {images/noisy KinectV2_crop}
%    \makebox[20pt]{\raisebox{72pt}{\rotatebox[origin=c]{90}{\nameref{para:PhoXi}}}}
%    \includegraphics[width=\dimexpr\linewidth-20pt\relax]
%    {images/noisy PhoXif_crop}
%    \caption{artificial range image with synthetic noise}
%    \label{fig:synthsL}
%\end{subfigure}\hfill
%\begin{subfigure}[t]{0.375\textwidth}
%    \includegraphics[width=\textwidth]  
%    {images/noisy cut KinectV1_crop}
%    \includegraphics[width=\textwidth]
%    {images/noisy cut KinectV2_crop}
%    \includegraphics[width=\textwidth]
%    {images/noisy cut PhoXif_crop}
%    \caption{cut out section}
%    \label{fig:synthsS}
%\end{subfigure}
%\caption{Artificial range images with added noise according to fitted models.}
%\label{fig:synths}
%\end{figure*}

\end{document}